\documentclass[10pt,twocolumn,letterpaper]{article}

\usepackage{ijcb}
\usepackage{times}
\usepackage{epsfig}
\usepackage{graphicx}
\usepackage{amsmath}
\usepackage{amssymb}
\usepackage{multirow}
\usepackage{subcaption}

\usepackage[norule,symbol,perpage]{footmisc}
\usepackage{hyperref}
\hypersetup{
    colorlinks=true,
    linkcolor=blue,
    filecolor=magenta,      
    urlcolor=cyan,
    pdftitle={Overleaf Example},
    pdfpagemode=FullScreen,
    }

\ijcbfinalcopy 

\ifijcbfinal\fi

\makeatletter
\def\ps@IEEEtitlepagestyle{
\def\@oddfoot{\mycopyrightnotice}
\def\@evenfoot{}
}
\def\mycopyrightnotice{
{\hfill \footnotesize 978-1-6654-3780-6/21/\$31.00 \copyright 2021 IEEE\hfill}
}
\makeatother

\begin{document}

\title{MixFaceNets: Extremely Efficient Face Recognition Networks}
\author{Fadi Boutros$^{1,2}$, Naser Damer$^{1,2}$, 
Meiling Fang$^{1,2}$, 
Florian Kirchbuchner$^{1}$, Arjan Kuijper$^{1,2}$\\
$^{1}$Fraunhofer Institute for Computer Graphics Research IGD,
Darmstadt, Germany\\
$^{2}$Mathematical and Applied Visual Computing, TU Darmstadt,
Darmstadt, Germany\\
 Email: {fadi.boutros@igd.fraunhofer.de}
}


\maketitle
\thispagestyle{empty}

\begin{abstract}
In this paper, we present a set of extremely efficient and high throughput models for accurate face verification, MixFaceNets which are inspired by Mixed Depthwise Convolutional Kernels. Extensive experiment evaluations on Label Face in the Wild (LFW), Age-DB, MegaFace, and IARPA Janus Benchmarks IJB-B and IJB-C datasets have shown the effectiveness of our MixFaceNets for applications requiring extremely low computational complexity. 
Under the same level of computation complexity ($\leq$ 500M FLOPs), our MixFaceNets outperform MobileFaceNets on all the evaluated datasets, achieving 99.60\% accuracy on LFW, 97.05\% accuracy on AgeDB-30, 93.60 TAR (at FAR1e-6) on MegaFace, 90.94 TAR (at FAR1e-4) on IJB-B and 93.08 TAR (at FAR1e-4) on IJB-C. With computational complexity between 500M and 1G FLOPs, our MixFaceNets achieved results comparable to the top-ranked models, while using significantly fewer FLOPs and less computation overhead, which proves the practical value of our proposed MixFaceNets. All training codes, pre-trained models, and training logs have been made available \small{ \url{https://github.com/fdbtrs/mixfacenets}}.
\end{abstract}


\vspace{-2mm}
\section{Introduction}
\label{sec:int}
\vspace{-1mm}

Recent advancements in the development of deep convolutional neural networks and the novelty of margin-based Softmax loss functions have significantly improved the State-Of-the-Art (SOTA) performance of face recognition to an extraordinary level \cite{deng2019arcface}, even under challenging conditions \cite{Damer_mask_ext_IET,DBLP:journals/corr/abs-2103-01716}.
However, deploying deep learning-based face recognition models in the embedded domains and other use-cases constrained by the computational capabilities and high throughput requirements is still challenging \cite{martinez2021benchmarking,DBLP:conf/iccvw/DengGZDLS19}.

Recently, a great progress in designing efficient face recognition solutions has been achieved by utilizing lightweight deep learning model architecture designed for common computer version tasks such as MobileNetV2 \cite{DBLP:conf/cvpr/SandlerHZZC18}, ShuffleNet  \cite{DBLP:conf/eccv/MaZZS18}, VarGNet \cite{DBLP:journals/corr/abs-1907-05653} for face recognition \cite{DBLP:conf/ccbr/ChenLGH18, DBLP:conf/iccvw/LiWHL19, DBLP:conf/iccvw/Martinez-DiazLV19,DBLP:conf/iccvw/YanZXZWS19}.
MobileFaceNet \cite{DBLP:conf/ccbr/ChenLGH18} was one of the earliest works that proposed an efficient face recognition model with around 1M of parameters and 439M FLOPs.
MobileFaceNet architecture is inspired by  MobileNetV2 \cite{DBLP:conf/cvpr/SandlerHZZC18}.
The architecture of MobileNetV2 is based on an inverted residual structure and the depthwise separable convolution \cite{DBLP:journals/corr/HowardZCKWWAA17}.
AirFace \cite{DBLP:conf/iccvw/LiWHL19}, ShuffleFaceNet \cite{DBLP:conf/iccvw/Martinez-DiazLV19} and VarGFaceNet \cite{DBLP:conf/iccvw/YanZXZWS19} model architectures are built from MobileNetV2 \cite{DBLP:conf/cvpr/SandlerHZZC18}, ShuffleNetV2 \cite{DBLP:conf/eccv/MaZZS18} and VarGNet \cite{DBLP:journals/corr/abs-1907-05653}, respectively, reaching high levels of accuracy using compact models with around 1G FLOPs computation complexity. ShuffleNetV2 utilizes channel shuffle operation proposed by ShuffleNetV1, achieving an acceptable trade-off between accuracy and computational efficiency. VarGNet \cite{DBLP:journals/corr/abs-1907-05653} proposed to fix the number of input channels in each group convolution instead of fixing the total group numbers in an effort to balance the computational intensity inside the convolutional block.
VarGFaceNet was the winner of the DeepGlint-Light track of ICCV Lightweight Face Recognition (LFR) Challenge (2019) \cite{DBLP:conf/iccvw/DengGZDLS19}. The deepglint-light track of the LFR challenge targets face recognition for environment constraint by the computational complexity of 1G FLOPs and memory footprint of 20M (around 5M of trainable parameters).
Very recently, MixNets proposed new convolution building block (MixConv) by extending vanilla depthwise convolution \cite{DBLP:journals/corr/HowardZCKWWAA17} with multiple kernel sizes. Using neural architecture search, MixNets developed highly efficient networks. MixNets outperformed previous mobile models including MobileNets \cite{DBLP:conf/cvpr/SandlerHZZC18,DBLP:journals/corr/HowardZCKWWAA17}, ShuffleNetV2 \cite{DBLP:conf/eccv/MaZZS18} on image classification and object detection tasks.  

In this work, we propose a set of extremely efficient architectures for accurate face verification and identification, namely the MixFaceNets. We opt to use MixNets as a baseline network structure to develop our MixFaceNets. We carefully designed tailored head and embedding settings that are suitable for face recognition.  We also extend the MixConv block with a channel shuffle operation aiming at increasing the discriminative ability of MixFaceNets. With computation complexity of 451M FLOPs, our MixFaceNet-S and ShuffleMixFaceNet-S achieved 99.60 and 99.56 \% accuracies on Labeled Faces in the Wild (LFW) \cite{LFWTech} and 92.23 and 93.60 TAR (at FAR1e-6) on MegaFace \cite{DBLP:conf/cvpr/Kemelmacher-Shlizerman16}  which are significantly higher than the ones achieved by MobileFaceNets \cite{DBLP:conf/ccbr/ChenLGH18} with a comparable level of computational complexity (99.55\% accuracy on LFW and 90.16\% TAR (at FAR1e-6) on MegaFace). Also, our MixFaceNets achieve comparable results to the SOTA solutions that have computation complexity of thousands of MFLOPs. 




\begin{figure*}[ht]
    \centering
    \includegraphics[width=0.79\textwidth]{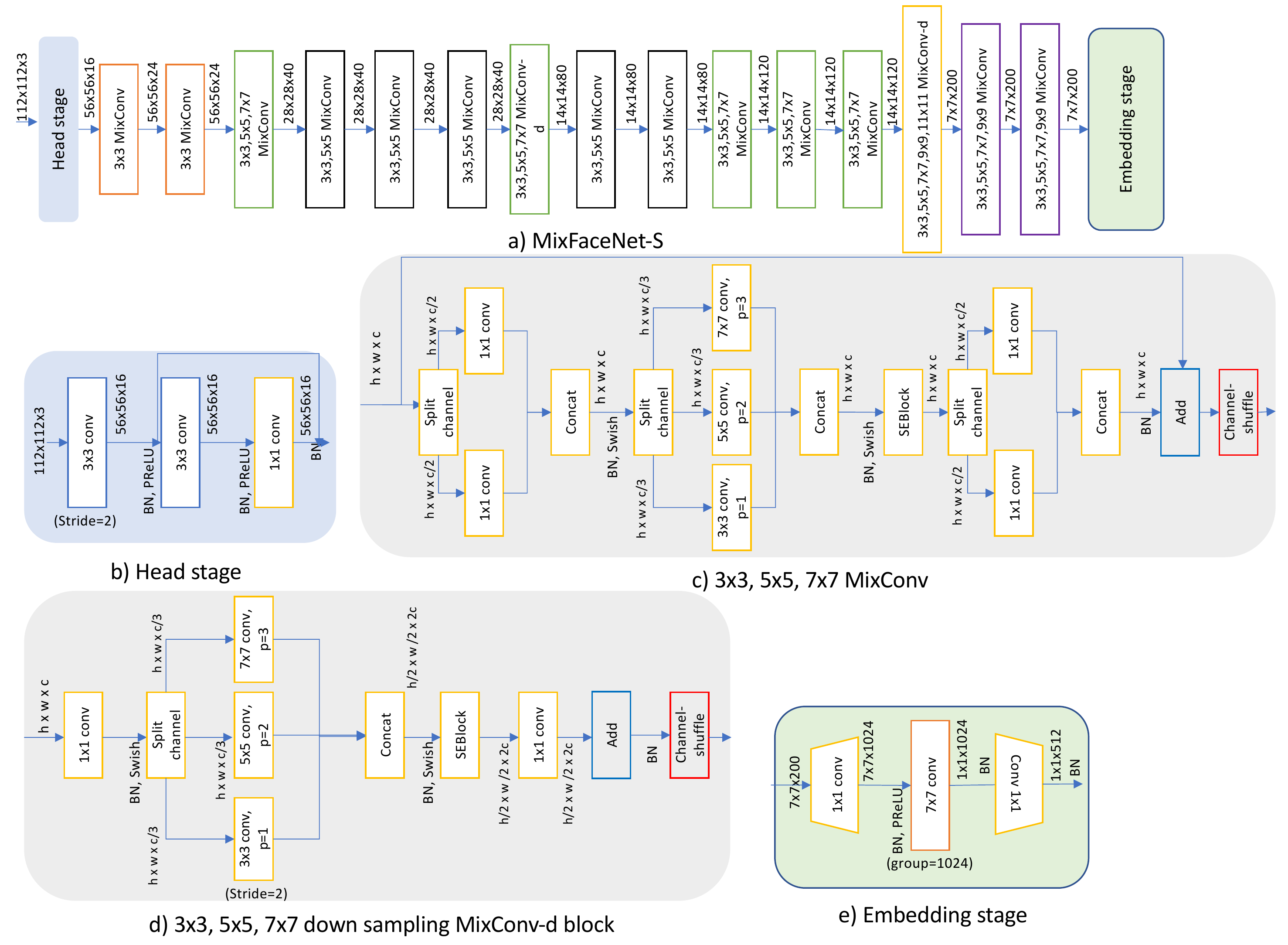}
        \vspace{-1mm}
    \caption{ An overview of our proposed MixFaceNet-S network architecture inspired by  MixNets \cite{DBLP:conf/bmvc/TanL19}. The input of MixFaceNet-S has size of $112 \times 112 \times 3$ and the output is a face embedding of dimension $512-d$. b) illustrates the head setting of MixFaceNet-S. We downsample the input on the first convolution (stride=2) then we add one residual block. d) is the MixConv block with multiple kernel sizes ($[(3,3,),(5,5),(7,7)]$) and channel shuffle operation. All MixConv blocks have the same structure as in (c) and the reduced blocks have the same structure as in (d).
    e) shows the embedding setting of MixFaceNet-S. We expand the channel from 200 to 1024, then we apply global depthwise convolution to obtain a $512-d$ embedding. The input and output size, kernel size, stride, and padding (\textit{p}) are shown for each convolution layer.
      }
    \label{fig:mixfacenet}
    \vspace{-4mm}
\end{figure*}

\vspace{-2mm}
\section{Related Works}
\label{sec:rw}
\vspace{-1mm}
Common deep learning-based face recognition models are typically computationally expensive for deployment on low computational power devices and high throughput processes. This challenge has received increased attention in the literature in the last few years \cite{martinez2021benchmarking,DBLP:conf/iccvw/DengGZDLS19}. In this section, we list out and discuss the recent efforts on designing an efficient deep learning model for face recognition. While there are a wide range of evaluation datasets for face recognition and each of the listed previous works reported the performance on a different set of evaluation datasets, we present in this section the reported accuracy by these previous works on the LFW \cite{LFWTech} as it is the most commonly reported benchmark. We also account for the computational complexity of the list of works in terms of the number of trainable parameters and the FLOPs when it is feasible.

The Light CNN \cite{DBLP:journals/tifs/WuHST18} was one of the earliest works that presented 3 network architectures for learning compact representation on a large-scale database.
The proposed architectures- Light CNN-4, Light CNN-9, and Light CNN-29, contain 4, 5.5, and 12.6m trainable parameters and FLOPs of 1.5G, 1, and 3.9G, respectively. The best performance among these three architectures on the LFW dataset (99.33\%) is achieved by the Light CNN-29. 
Compared to the recent efficient face recognition models \cite{martinez2021benchmarking},  Light CNN architecture is considered computational expensive. ShiftNet \cite{DBLP:conf/cvpr/WuWYJZGGGK18} proposed shift-based modules as an alternative to spatial convolutions and it is adopted for the face recognition task. The presented face model contains 0.78m parameters and it achieved 96\% accuracy on the LFW dataset.
MobileFaceNets \cite{DBLP:conf/ccbr/ChenLGH18} is based on MobileNetV2 \cite{DBLP:conf/cvpr/SandlerHZZC18} and it achieved a high accuracy on LFW dataset with around one million of trainable parameter and 443M FLOPs.   
MobileFaceNets model architecture is based on the residual bottlenecks proposed by MobileNetV2 \cite{DBLP:conf/cvpr/SandlerHZZC18} and depth-wise separable convolutions layer, which allows building CNN with a smaller set of parameters compared to standard CNNs. Different from MobielNetV2 architecture, MobileFaceNet uses Parametric Rectified Linear Unit (PReLU) \cite{DBLP:conf/iccv/HeZRS15} as the non-linearity in all convolutional layers and replaces the last global average pooling with linear global depth-wise convolution layer as a feature output layer.
The best-reported accuracy on LFW by MobileFaceNets (1m parameters and 433M FLOPs) was 99.55\%.

Several recent works utilized lightweight models designed for a common computer vision task to learn face representation \cite{DBLP:conf/iccvw/Martinez-DiazLV19,DBLP:conf/iccvw/YanZXZWS19}. ShuffleFaceNet \cite{DBLP:conf/iccvw/Martinez-DiazLV19} is a compact face recognition model based on ShuffleNet \cite{DBLP:conf/eccv/MaZZS18}. Similar to MobileFaceNet \cite{DBLP:conf/ccbr/ChenLGH18}, ShuffelFaceNet replaces the last global average pooling layer with a global depth-wise convolution layer and a Rectified Linear Unit (ReLU) with PReLU. ShuffleFaceNet presented four models with different complexity levels. The best-reported accuracy on LFW was 99.67\% (2.6M parameters and FLOPs of 577.5M).  VarGFaceNet \cite{DBLP:conf/iccvw/YanZXZWS19} deployed variable group convolutional network proposed by VarGNet \cite{DBLP:journals/corr/abs-1907-05653} to design a compact face recognition model with 5m trainable parameters and 1G FLOPS. VarGFaceNet adds squeeze and excitation (SE) block on the VarGNet block, replaces ReLU with PReLU, and uses variable group convolution along with pointwise convolution as the feature output layer. VarGFaceNet achieved 99.683 \% accuracy on LFW. This accuracy is increased to 99.85 \% by training VarGFaceNet with recursive knowledge distillation. 
However, the computational cost of VarGFaceNet is higher than the ones of ShuffelFaceNet and MobileFaceNet.
AirFace \cite{DBLP:conf/iccvw/LiWHL19} proposed to increase the MobileFaceNet network width and the depth and adding attention module, achieving 99.27\% accuracy on LFW.
The work also presented a loss function named Li-ArcFace which is based on ArcFace. Li-ArcFace demonstrates better converging and performance than ArcFace loss on low dimensional features embedding.
The proposed model by AirFac has a computational cost of 1G FLOPS.

In a recent survey by Martinez‑Diaz et al. \cite{martinez2021benchmarking}, the computational requirements and the verification performance of five lightweight model architectures are analyzed and evaluated. The evaluated models are MobileFaceNet (0.9G FLOPs and 2.0m parameters), VarGFaceNet \cite{DBLP:conf/iccvw/YanZXZWS19} (1G FLOPs and 5m parameters), ShufeFaceNet \cite{DBLP:conf/iccvw/Martinez-DiazLV19} (577.5M FLOPs and 2.6m parameters), MobileFaceNetV1 (1.1G FLOPs and 3.4m parameters), and ProxylessFaceNAS (0.9G FLOPs and 3.2m parameters). 
MobileFaceNetV1, ProxylessFaceNAS and MobileFaceNet are extended versions of MobileNetV1 \cite{DBLP:journals/corr/HowardZCKWWAA17}, ProxylessNAS \cite{DBLP:conf/iclr/CaiZH19}, and MobileFaceNets \cite{DBLP:conf/ccbr/ChenLGH18}, respectively.
The evaluated models in  \cite{martinez2021benchmarking}- ShufeFaceNet, VarGFaceNet, MobileFaceNet achieved very close accuracy on LFW (99.7\%) and other evaluated datasets, while MobileFaceNetV1 and ProxylessFaceNAS achieved slightly lower accuracy.

Among the previous listed works, MobileFaceNets \cite{DBLP:conf/ccbr/ChenLGH18} is the only architecture that achieved high accuracy with less than 500M FLOPs. 
With the almost same number of FLOPs as in MobileFaceNets, our MixFaceNets outperform  MobileFaceNets and achieved competitive results to other models with fewer FLOPs using extremely efficient architecture as shown in Tables \ref{tab:lfw} and \ref{tab:megaface}.

\vspace{-2mm}
\section{Approach}
\label{sec:meh}
\vspace{-1mm}
This section presents the architecture of our efficient MixFaceNets designed for accurate face verification. Figure \ref{fig:mixfacenet} illustrates the architecture of the MixFaceNet, partially inspired by MixNets \cite{DBLP:conf/bmvc/TanL19}.  To improve the accuracy and the discriminative ability of MixNet, we (a) implement different head settings, (b) introduce channel shuffle operation to the MixConv block, and (c) propose different embedding settings.
We discuss in this section the MixConv as an inspiration to our work. Then, we present the detailed architecture of our MixFaceNet.

\vspace{-1mm}
\subsection{Mixed Depthwise Convolutional Kernels}
\vspace{-1mm}
Depthwise Convolution is one of the most popular building block for mobile models \cite{DBLP:journals/corr/HowardZCKWWAA17,DBLP:conf/cvpr/SandlerHZZC18,DBLP:conf/eccv/MaZZS18}.
Depthwise Convolutional applies a single convolution filter over each channel of input, thus it reduces the number of parameters and achieves computational efficiency while maintaining the discriminative ability of the convolution
\cite{DBLP:journals/corr/HowardZCKWWAA17}.
Mixed Depthwise Convolutional Kernels (MixConv) \cite{DBLP:conf/bmvc/TanL19} extends
vanilla depthwise convolution by using multiple kernel sizes in a single convolution. MixConv depends on mixing up multiple kernel sizes in a single convolution by splitting up the input of convolution into groups and applying different kernel sizes to each group. 
Different from vanilla depthwise convolution, MixConv can capture different patterns from convolution input at various resolutions. Also, it requires fewer parameters and it is more computationally efficient than using a single kernel e.g. using multiple kernels of size $[(3,3),(5,5),(7,7)]$ is more computation efficient than using a single kernel of size $7 \times 7$.
For example, given a convolution input of size $w \times h \times c$ and multiple kernels of size $[(3,3),(5,5),(7,7)]$, MixConv split the input into 3 groups, each of them has a dimension of $w \times h \times c/3$. Then it uses different kernels for each of these groups. Finally, the three outputs are concatenated to produce the final convolution output. An example of MixConv and downsample-MicConv blocks are shown as part of Figure \ref{fig:mixfacenet}.
Unlike manually designed mobile models \cite{DBLP:conf/eccv/MaZZS18,DBLP:conf/iccvw/Martinez-DiazLV19,DBLP:conf/cvpr/SandlerHZZC18},
MixConv utilized neural architecture search to develop new series of MixConv-based networks, namely MixNets (MixNet-S, MixNet-M, and MixNet-L).  MixNet-S and MixNet-M are developed using neural architecture search, while MixNet-L is obtained by scaling up the number of channels in each block by a factor of 1.3. For details about the network structure and search space, we refer to the original work \cite{DBLP:conf/bmvc/TanL19}.

\subsection{MixFaceNet Architecture}
\vspace{-1mm}
We deploy MixNets \cite{DBLP:conf/bmvc/TanL19} as baseline network structure to develop our proposed MixFaceNets. 
Figure \ref{fig:mixfacenet} illustrates the network architecture for MixFaceNet-S. For the network head, we apply fast down-sampling in the first $3 \times 3$ convolution (stride=2) followed by batch normalization \cite{DBLP:conf/icml/IoffeS15} and PReLU non-linearity \cite{DBLP:conf/iccv/HeZRS15}. 
Then, we use one residual block as shown in Figure \ref{fig:mixfacenet}.b.
For the MixFaceNet-S network, we use the same global structure as MixNet-S. However, different from MixNet-S, we did not apply down-sampling at the first convolution after the head stage to reserve as much as possible information at the earliest stage of the network. 
We propose to mix up both channels and kernels to increase the discriminative ability of MixFaceNet and improve the model accuracy. We achieve that by introducing shuffle operation to the MixConv block.
The channel shuffle operation is proposed by \cite{DBLP:conf/iccvw/Martinez-DiazLV19} to enabling information flowing between different groups of channels. We apply a channel shuffle operation with a group value of 2 after each MixConv block.
Thus, MixFaceNet can capture high and low-resolution patterns at different scales and it also enables information communication between various groups of channels.
Figures \ref{fig:mixfacenet}.c and \ref{fig:mixfacenet}.d show the detailed structure of MixConv and downsampling MixConv blocks  with the channel shuffle operation.
All MiXConv blocks uses swish as an activation function \cite{DBLP:conf/iclr/RamachandranZL18} followed by batch normalization. MixConv includes also squeeze-and-excitation (SEBlock) \cite{DBLP:conf/cvpr/HuSS18} at the end of each block.
Finally, to obtain the feature embedding of the input face image, we replace the last global average pooling layer with global depth-wise convolution as presented in the next section. We propose 3 network architectures, MixFaceNet-XS, MixFaceNet-S, and MixFaceNet-M.   MixFaceNet-S network architecture is illustrated in Figure \ref{fig:mixfacenet}.a. 
MixFaceNet-XS is obtained by scaling up MixFaceNet-S with a depth multiplier of 0.5. MixFaceNet-M has the same global network architecture as MixNet-M \cite{DBLP:conf/bmvc/TanL19} with the same strategies applied to MixFaceNet-S i.e. head setting and embedding settings. All proposed MixFaceNets are trained and evaluated with/without channel shuffle operation. The models trained with channel shuffle operation will be noted as ShuffleMixFaceNet-XS, ShuffleMixFaceNet-S, and ShuffleMixFaceNet-M.

\vspace{-2mm}
\subsubsection{Embedding Setting}
\vspace{-1mm}
MixNets use global average pooling before the classification layer as a feature output layer. This is common choice for most of the classical compact deep learning models \cite{DBLP:journals/corr/HowardZCKWWAA17,DBLP:conf/cvpr/SandlerHZZC18,DBLP:conf/eccv/MaZZS18}. However, previous works in \cite{deng2019arcface,DBLP:conf/ccbr/ChenLGH18} observed that CNNs with a fully connected layer (FC) or global depthwise convolution are more accurate than the ones with global average pooling for face recognition and verification. A fully connected layer has been used in many of the recent deep face recognition models to obtain face representations \cite{deng2019arcface}. However, using FC on top of the last convolutional layer will add a large number of parameters to the model. And thus, it extremely increases the memory footprint and reduces the throughput.  For example, giving the last convolutional layer of CNN with a kernel size of $7 \times 7$ (as in MixNet) and output feature maps of size $200$, the output of this layer, in this case, has a size of $7 \times 7 \times 200$. Using FC of size $512-d$ on top of the previous layer will add additional 5M parameters to the network ($7 \times 7 \times 200 \times 512$). Even for small FC, $128-d$, the number of additional parameters causes by FC will be 1.2M. Thus, using FC is not the optimal choice for an efficient face recognition model. 
Using global depthwise convolution is a common choice for most of the previous works proposing efficient face recognition models as it contains fewer parameters than FC and it can lead to higher verification performance than using global average pooling \cite{DBLP:conf/ccbr/ChenLGH18}. 
Therefore, we replace the global average pooling with global depthwise convolution. Specifically, we first add $1 \times 1$ convolutional layer (Conv1) with stride=1 and zero paddings followed by batch normalization \cite{DBLP:conf/icml/IoffeS15} and PReLU none-linearity \cite{DBLP:conf/iccv/HeZRS15}. In Conv1, we expand the channel from 200  to 1024. Then, we use $7 \times 7$ convolution layer (stride=1, padding=0 and grouping=1024) followed by batch normalization.  Finally, we use $1 \times 1$ convolution with 512 output channels followed by batch normalization to obtain the final feature embedding which is of size $512-d$, as shown in Figure \ref{fig:mixfacenet}.e.
\begin{table*}[ht!]
\centering
\begin{tabular}{|c|c|l|c|c|}
\hline
Method                      & FLOPs (M) & \# Params. (M) & LFW  (\%) & AgeDB-30 (\%) \\ \hline
ArcFace (LResNet100E-IR) \cite{deng2019arcface}    & 24211     & 65.2          & 99.83     & 98.15         \\ \hline \hline
AirFace     \cite{DBLP:conf/iccvw/LiWHL19}                & 1000      & -              & 99.27    & -             \\
ShuffleFaceNet 2× \cite{DBLP:conf/iccvw/Martinez-DiazLV19}          & 1050      & 4.5           & 99.62     & 97.28         \\
ShuffleFaceNet 1.5×   \cite{DBLP:conf/iccvw/Martinez-DiazLV19}      & 577.5     & 2.6           & 99.67     & 97.32         \\
VarGFaceNet \cite{DBLP:conf/iccvw/YanZXZWS19}                 & 1022      & 5         & 99.68     &  98.10     \\
MobileFaceNet \cite{martinez2021benchmarking}              & 933       & 2.0          & 99.7      & 97.6          \\
MobileFaceNetV1 \cite{martinez2021benchmarking}             & 1100      & 3.4          & 99.4      & 96.4          \\
ProxylessFaceNAS \cite{martinez2021benchmarking}            & 900       & 3.2          & 99.2      & 94.4          \\
MixFaceNet-M (ours)         & 626.1     & 3.95          & 99.68     & 97.05         \\
ShuffleMixFaceNet-M (ours)  & 626.1     & 3.95         & 99.60      & 96.98         \\ \hline \hline
MobileFaceNets   \cite{DBLP:conf/ccbr/ChenLGH18}           & 439.8     & 0.99             & 99.55     & 96.07         \\
ShuffleFaceNet 0.5×   \cite{DBLP:conf/iccvw/Martinez-DiazLV19}      & 66.9      & 0.5           & 99.23     & 93.22         \\
MixFaceNet-S (ours)         & 451.7     & 3.07         & 99.60    & 96.63         \\
ShuffleMixFaceNet-S (ours)  & 451.7     & 3.07         & 99.58     & 97.05        \\
MixFaceNet-XS  (ours)       & 161.9     & 1.04          &99.60     & 95.85         \\
ShuffleMixFaceNet-XS (ours) & 161.9     & 1.04         & 99.53     & 95.62         \\ \hline
\end{tabular}%
\caption{MixFaceNets verification accuracies on LFW and AgeDB-30 datasets. The first row of the table show the achieved result by the current SOTA ReNet100 models. The table is divided into two parts. The first part of the table shows the achieved result by models that have computational complexity between 500 and 1000M FLOPs. The second part of the table shows the achieved by models that have computational complexity less than 500M FLOPs. The number of decimal points is reported as in the related works. }
\label{tab:lfw}
\vspace{-3mm}
\end{table*}

\vspace{-2mm}
\section{Experimental Setup} 
\label{sec:es}
\vspace{-1mm}
\paragraph{Dataset:}
We use the MS1MV2 dataset \cite{deng2019arcface} to train our MixFaceNet models.
The MS1MV2 is a refined version of the MS-Celeb-1M \cite{DBLP:conf/eccv/GuoZHHG16} by \cite{deng2019arcface} and it contains 5.8M images of 85K identities. The Multi-task Cascaded Convolutional Networks (MTCNN) solution \cite{zhang2016joint} is used to detect and align face images.  The MixFaceNet models process aligned and cropped face images of the size $112 \times 112 \times 3$ to produce $512-d$ feature embeddings. 
We evaluate our MixFaceNets on the widely used LFW \cite{LFWTech} and on the AgeDB-30 \cite{DBLP:conf/cvpr/MoschoglouPSDKZ17} datasets. Also, we report the performance of the MixFaceNets on large scale evaluation datasets including MegaFace \cite{DBLP:conf/cvpr/Kemelmacher-Shlizerman16}, IJB-B \cite{DBLP:conf/cvpr/WhitelamTBMAMKJ17} and IJB-C \cite{DBLP:conf/icb/MazeADKMO0NACG18}.

\vspace{-2mm}
\paragraph{MixFaceNets Training Setup:}
The proposed models in this paper are implemented using Pytorch. All models are trained using ArcFace loss \cite{deng2019arcface}. We set the margin value of ArcFace loss to 0.5 and the feature scale to 64.
We set the batch size to 512 and train our model using distributed Partial-FC algorithm \cite{an2020partical_fc} on one machine with 4 Nvidia GeForce RTX 6000 GPUs to enable faster training on a single node. All models are trained with Stochastic Gradient Descent (SGD) optimizer with an initial learning rate of 1e-1. 
We set the momentum to 0.9 and the weight decay to 5e-4. The learning rate is divided by 10 at 80k, 140k, 210k, and 280k training iterations. During the training, we evaluate the model on LFW and AgeDB after each 5650 training iterations. 
The training is stopped after 300k iterations. During the testing, the classification layer is removed and the feature is extracted from the last layer, which is of the size $512-d$. We used euclidean distance between feature vectors in all experiments for comparison.

\begin{table*}[]
\centering
\resizebox{\textwidth}{!}{%
\begin{tabular}{|c|c|c|c|c|c|c|c|c|}
\hline
\multirow{2}{*}{Method} & \multirow{2}{*}{MFLOPs}  & \multirow{2}{*}{Params (M)}  & \multicolumn{2}{c|}{MegaFace} & \multicolumn{2}{c|}{MegaFace (R)} & \multicolumn{2}{c|}{IJB} \\ \cline{4-9} 
                         &          &        & Rank-1 (\%) & TAR at FAR1e–6 & Rank-1  (\%)  & TAR at FAR1e–6  & IJB-B & IJB-C \\ \hline
ArcFace (LResNet100E-IR) \cite{deng2019arcface}  & 24211  & 65.2  & 81.03  & 96.98          & 98.35      & 98.48              & 94.2  & 95.6  \\ \hline \hline
AirFace   \cite{DBLP:conf/iccvw/LiWHL19}             & 1000       & -      &  80.80 & 96.52         & 98.04      & 97.93             & -     & -     \\
MobileFaceNet     \cite{martinez2021benchmarking}       & 933     & 2.0  & 79.3   & 95.2           & 95.8       & 96.8               & 92.8  & 94.7  \\
ShuffleFaceNet   \cite{martinez2021benchmarking,DBLP:conf/iccvw/Martinez-DiazLV19}         & 577.5     & 2.6   & 77.4   & 93.0           & 94.1       & 94.6               & 92.3  & 94.3  \\
MobileFaceNetV1 \cite{martinez2021benchmarking}         & 1100     & 3.4   & 76.0   & 91.3           & 91.7       & 93.0               & 92.0  & 93.9  \\
VarGFaceNet   \cite{martinez2021benchmarking,DBLP:conf/iccvw/YanZXZWS19}           & 1022     & 5.0   & 78.20   & 93.9           & 94.9       & 95.6               &92.9  &94.7 \\
ProxylessFaceNAS  \cite{martinez2021benchmarking}       & 900     & 3.2  & 69.7   & 82.8           & 82.1       & 84.8               & 87.1  & 89.7  \\
MixFaceNet-M   (ours)          & 626.1   & 3.95  & 78.2   & 94.26          & 94.95      & 95.83              & 91.55 & 93.42 \\
ShuffleMixFaceNet-M (ours)      & 626.1  & 3.95  & 78.13  & 94.24          & 94.64      & 95.22              & 91.47 & 93.5  \\ \hline \hline
MobileFaceNets \cite{DBLP:conf/ccbr/ChenLGH18}            & 439.8   & 0.99    & -      & 90.16          & -          & 92.59              & -     & -     \\
MixFaceNet-S (ours)            & 451.7    & 3.07  & 76.49  & 92.23          & 92.67      & 93.79              & 90.17 & 92.30 \\
ShuffleMixFaceNet-S (ours)     & 451.7    & 3.07  & 77.41  &93.60      & 94.07    & 95.19              & 90.94 & 93.08 \\
MixFaceNet-XS            & 161.9   & 1.04  & 74.18  & 89.40          & 89.35      & 91.04              & 88.48 & 90.73 \\
ShuffleMixFaceNet-XS (ours)    & 161.9   & 1.04 & 73.85  & 89.24          & 88.823     & 91.03              & 87.86 & 90.43 \\ \hline
\end{tabular}%
}
\vspace{-1mm}
\caption{ The achieved results on large-scale evaluation datasets- MegaFace, IJB-B, and IJB-C.  
The results on MegaFace and MegaFace (R) \cite{deng2019arcface} using FaceScrube as probe set are reported as face identification (Rank-1 \%) and verification (TAR at FAR1e–6) for different lightweight models.  The last two columns of the table show 1:1 verification TAR (at FAR=1e-4) on IJB-B and IJB-C.   The first row reports the evaluation result using SOTA face recognition model- ArcFace (LResNet100E-IR) which contains 65.2M parameters and 24211M FLOPs. The rest of the table is organized in two parts- models with computational complexity between 500 and 1000M FLOPs and models with less than 500M FLOPs. The number of decimal points is reported as in the related works.
Considering the computation complexity, our MixFaceNet models are evaluated as ones of top-ranked models.}
\label{tab:megaface}
\vspace{-4mm}
\end{table*}

\vspace{-1mm}
\section{Results}
\label{sec:res}
\vspace{-1mm}

This section presents the achieved result by the MixFaceNets on different benchmarks.
We acknowledge the evaluation metrics in the ISO/IEC 19795-1 \cite{mansfield2006information} standard. However, for the sake of comparability and reproducibility, we follow the evaluation metrics used in the utilized benchmarks and the previous works reporting on them.

\vspace{-1mm}
\subsection{Result on LFW and AgeDB-30}
\vspace{-1mm}

\begin{figure*}[ht]
     \centering
     \begin{subfigure}[b]{0.33\textwidth}
         \centering
         \includegraphics[width=\textwidth]{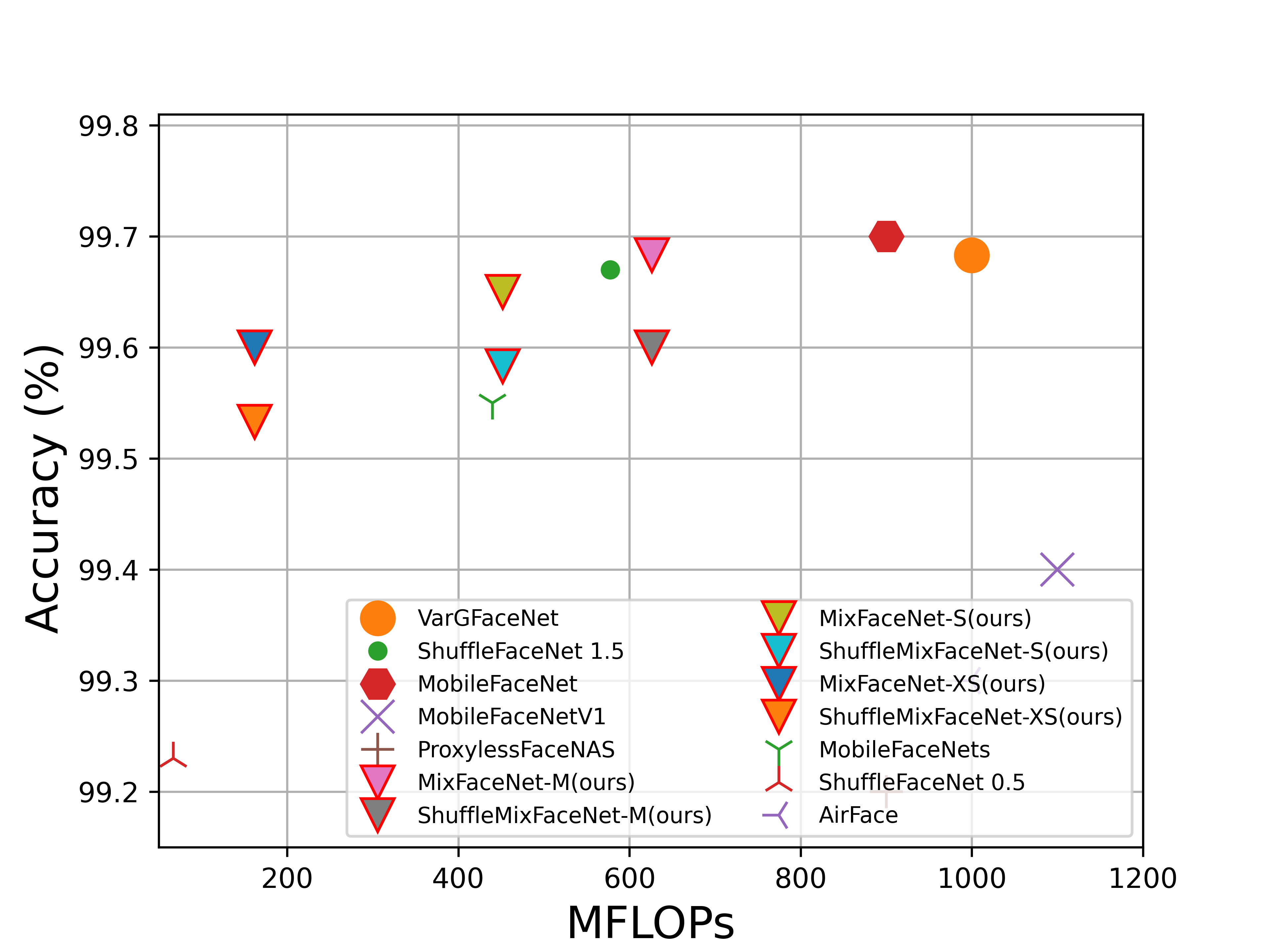}
         \caption{LFW}
         \label{fig:lfw}
     \end{subfigure}
      \begin{subfigure}[b]{0.33\textwidth}
         \centering
         \includegraphics[width=\textwidth]{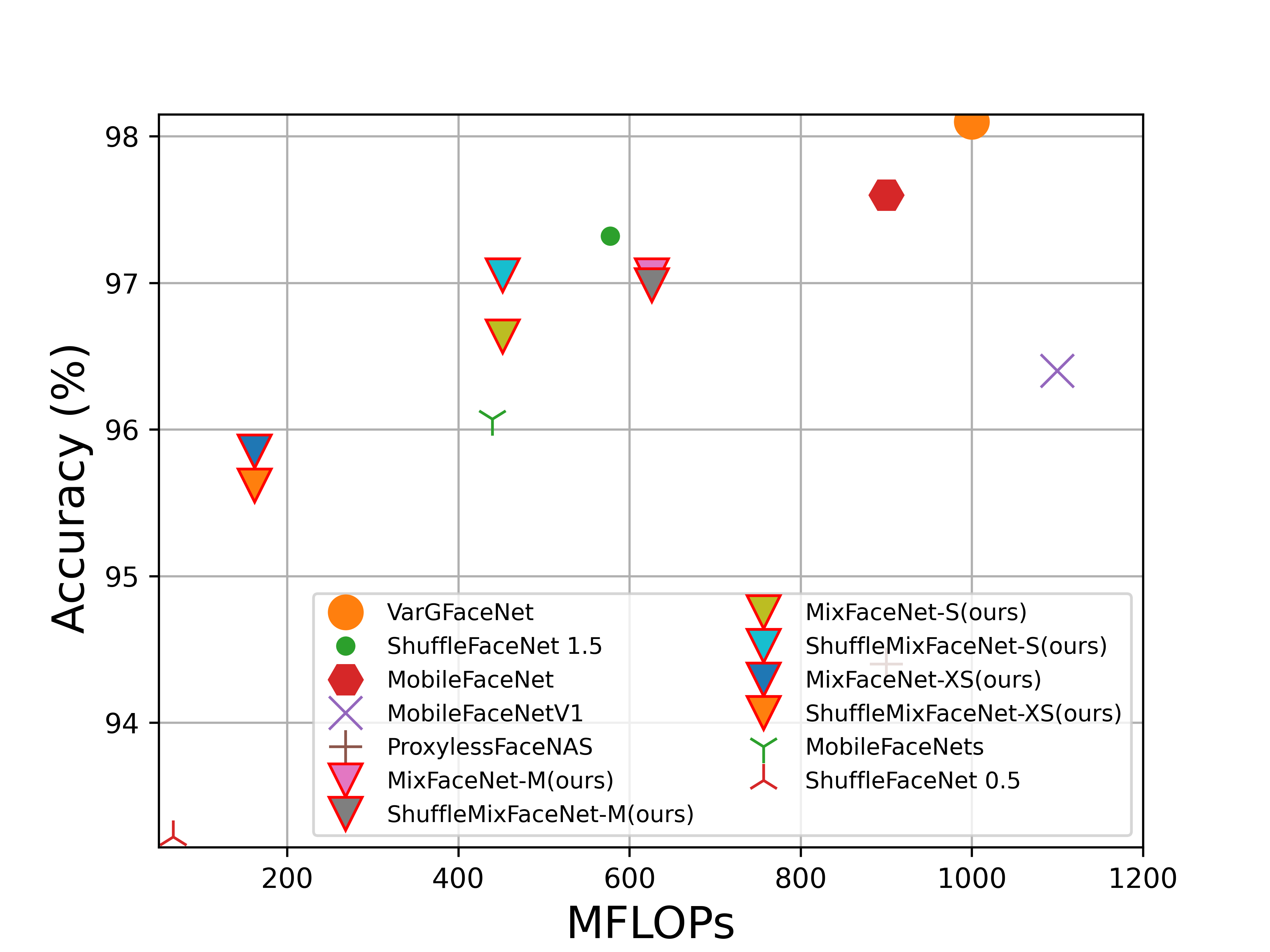}
         \caption{AgeDB-30}
         \label{fig:agedb}
     \end{subfigure}
           \begin{subfigure}[b]{0.33\textwidth}
         \centering
         \includegraphics[width=\textwidth]{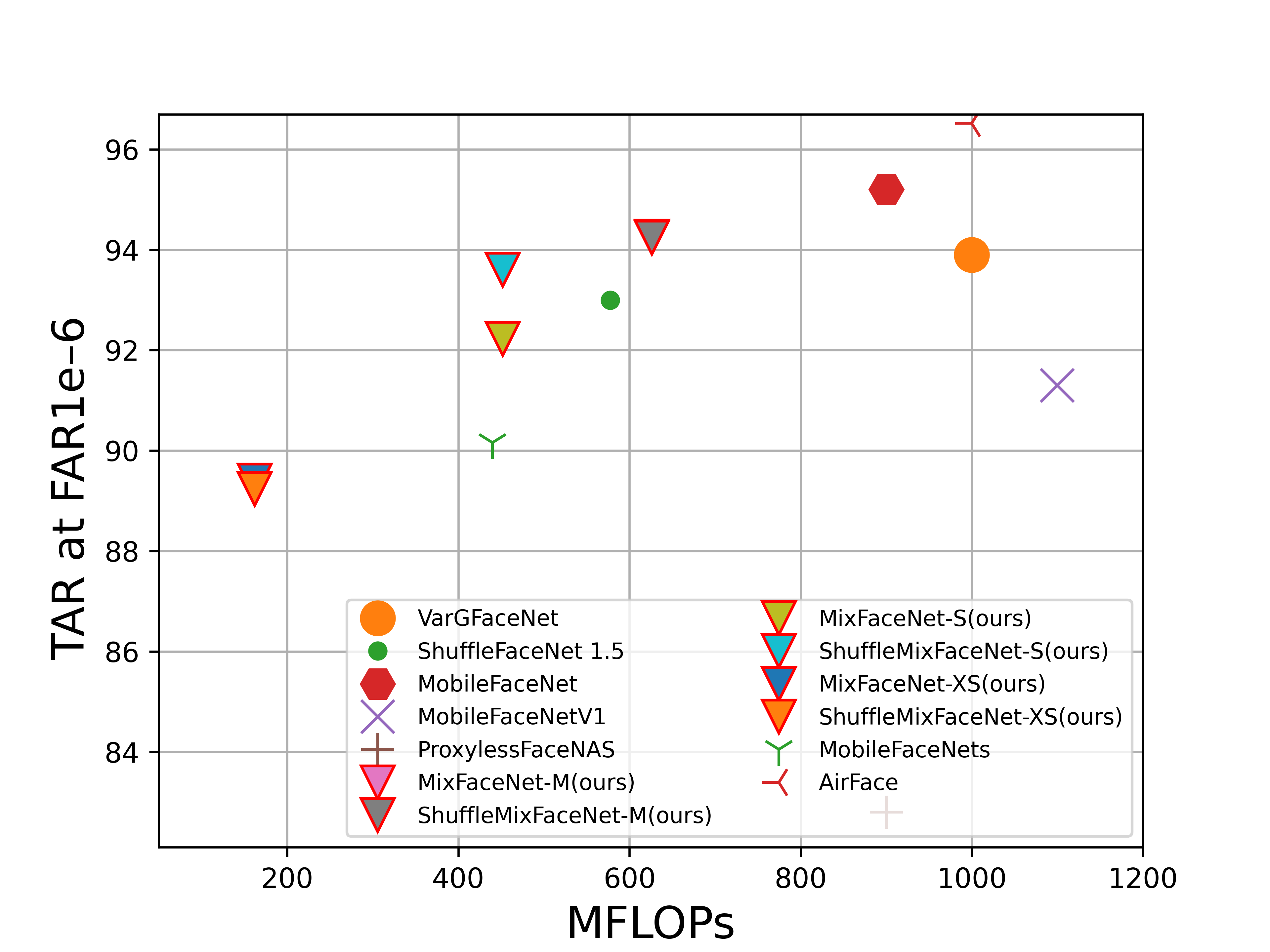}
         \caption{MegaFace}
         \label{fig:megaface}
     \end{subfigure}
     
          \vspace{-1mm}
    \begin{subfigure}[b]{0.33\textwidth}
         \centering
         \includegraphics[width=\textwidth]{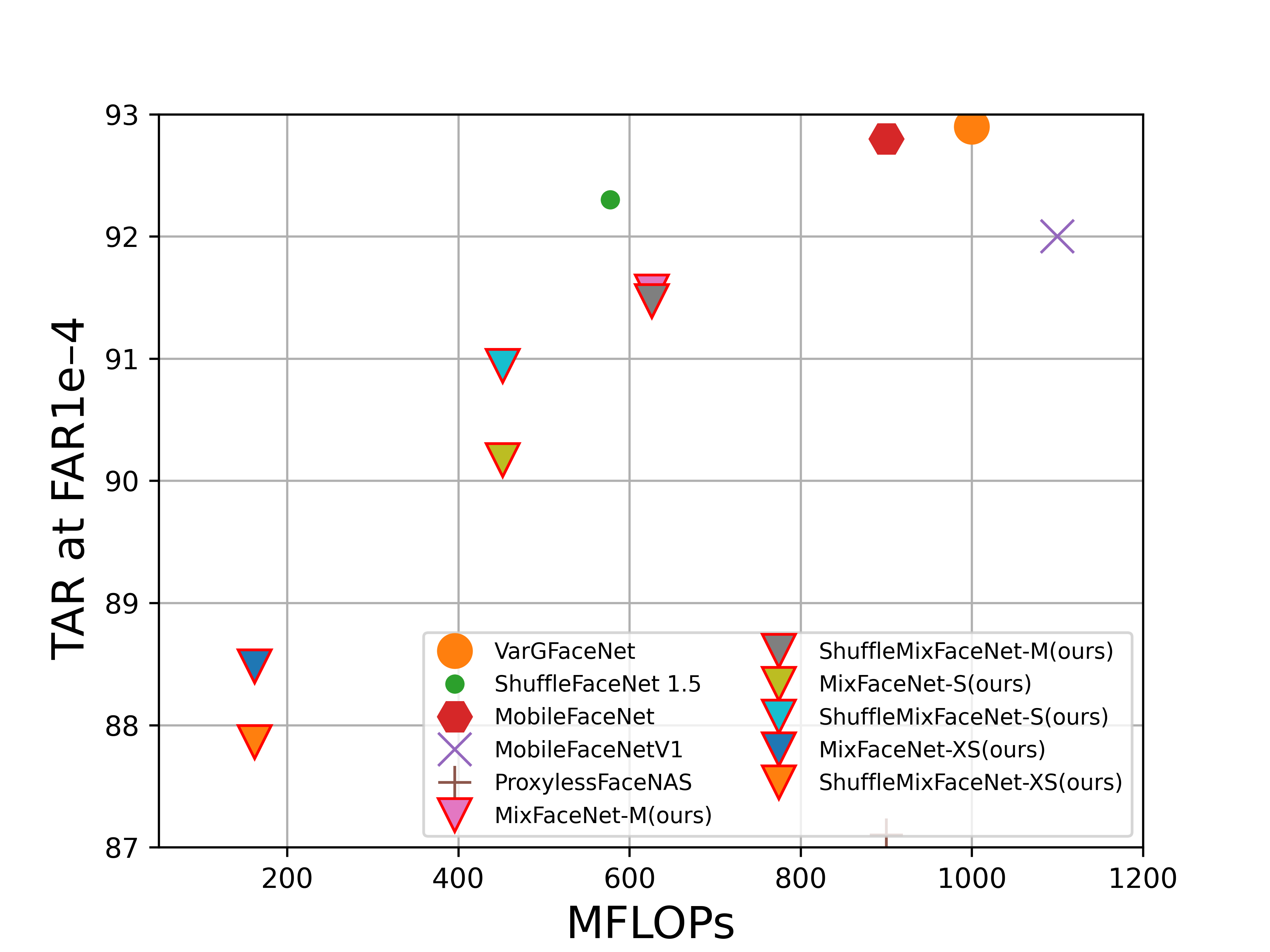}
         \caption{IJB-B}
         \label{fig:ijbb}
     \end{subfigure}
      \begin{subfigure}[b]{0.33\textwidth}
         \centering
         \includegraphics[width=\textwidth]{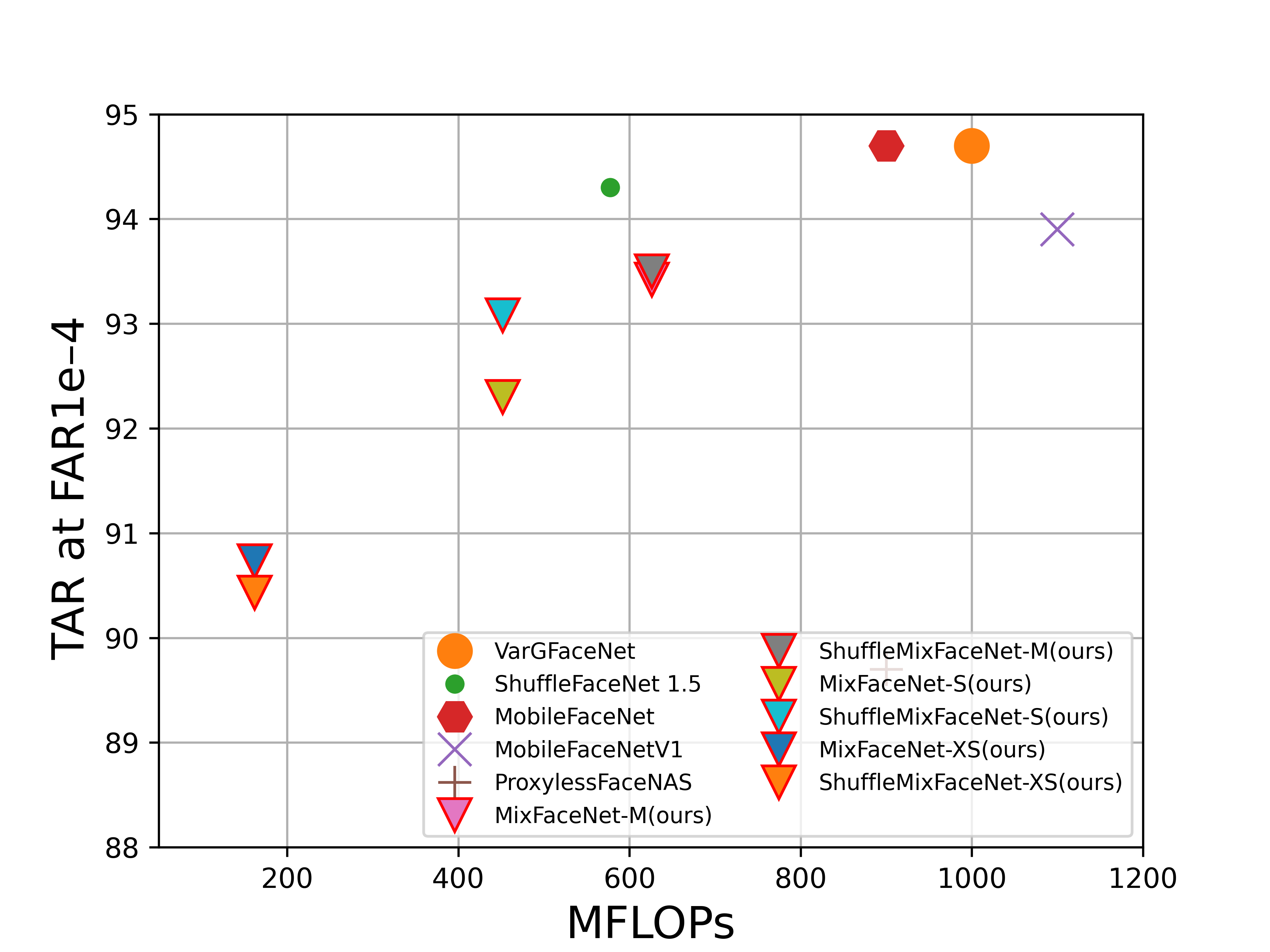}
         \caption{IJB-C}
         \label{fig:ijbc}
     \end{subfigure}
           \begin{subfigure}[b]{0.33\textwidth}
         \centering
         \includegraphics[width=\textwidth]{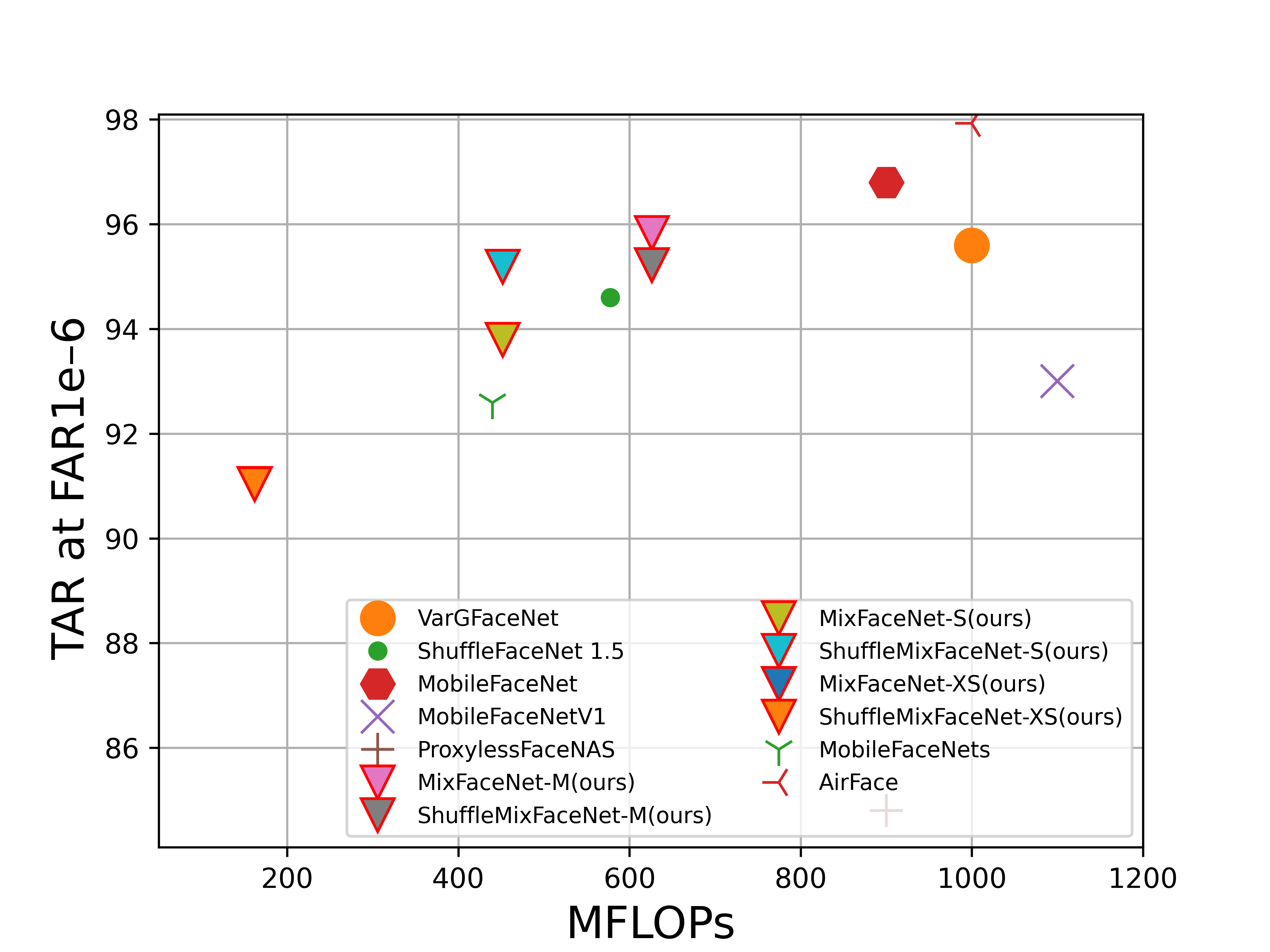}
         \caption{MegaFace (R)}
         \label{fig:megafacer}
     \end{subfigure}
     \vspace{-5mm}
        \caption{  FLOPs vs. performance on LFW (accuracy), AgeDB-30 (accuracy),  MegaFace (TAR at FAR1e-6), IJB-B (TAR at FAR1e-4),  IJB-C (TAR at FAR1e-4) and refined version of MegaFace, noted as MegaFace (R), (TAR at FAR1e-6).
        Our MixFaceNet models are highlighted with triangle marker and red edge color.
       }
        \label{fig:flops}
\vspace{-3mm}
\end{figure*}

LFW \cite{LFWTech} is one of the widely used datasets for unconstrained face verification. The dataset contains 13,233 images of 5749 different identities. The result on LFW is reported as verification accuracy (as defined in \cite{LFWTech}) following the unrestricted with labeled outside data protocol using the standard 6000 comparison pairs defined in \cite{LFWTech}. AgeDB \cite{DBLP:conf/cvpr/MoschoglouPSDKZ17} is common used in-the-wild dataset for evaluating age-invariant face verification. It contains 16,488 images of 568 different identities. We report the performance as verification accuracy for AgeDB-30 (years gap 30) as it is the most challenging subset of AgeDB. Also, it is the commonly reported set of AgeDB by the recent SOTA face recognition models. Similar to the LFW, we followed the standard protocol provided by AgeDB to evaluate our models on AgeDB-30 \cite{DBLP:conf/cvpr/MoschoglouPSDKZ17}. Table \ref{tab:lfw} shows the achieved result on LFW and AgeDB-30. 
We first report the result for one of the top-ranked face recognition models, ArcFace (LResNet100E-IR) \cite{deng2019arcface}, to give an indication of the current SOTA performance on LFW (99.83 \%) and AgeDB-30(98.15\%). Although, the ArcFace (LResNet100E-IR) model \cite{deng2019arcface} is far from being considered an efficient model, in comparison to lightweight models, with 24211M FLOPs and 65.2m parameters. 
Then, the second section of Table \ref{tab:lfw} presents the achieved result by the recent lightweight models that have computational complexity between 500 and 1000M FLOPs.
The best-reported result on LFW (99.70\% accuracy) is achieved by the MobileFaceNet \cite{martinez2021benchmarking} (933M FLOPs). 
Our MixFaceNet-M achieved a competitive result on LFW (99.68\% accuracy) using 38\% fewer FLOPs (626M).
A similar result has been achieved on AgeDB-30. Our MixFaceNets achieved very close accuracy to the current SOTA models using a more efficient model architecture with almost the same number of parameters.
Among all models that have computational complexity less than 500M FLOPs, our MixFaseNet models outperform all listed models including MobileFaceNets \cite{DBLP:conf/ccbr/ChenLGH18} on LFW and AgeDB-30. 
Similar conclusion can be seen in the Figure \ref{fig:lfw} and \ref{fig:agedb}. It can be clearly noticed that our MixFaceNet achieved the highest accuracies on LFW and AgeDB-30 when considering the same level of computational complexity.

\vspace{-1mm}
\subsection{Result on IJB-B and IJB-C}
\vspace{-1mm}

The IARPA Janus Benchmark-B ((IJB-B) face dataset  consists of 1,845 subjects of 21,798 still images and 55,026 frames from 7,011 videos \cite{DBLP:conf/cvpr/WhitelamTBMAMKJ17}. The IJB-B verification protocol provides a list of 10,270 genuine comparisons and 8M impostor comparisons.
The IARPA Janus Benchmark–C (IJB-C) face dataset is an extension of IJB-B by increasing the database variability and size with additional 1,661 new subjects \cite{DBLP:conf/icb/MazeADKMO0NACG18}.  The IJB-C consists of 31,334 still images and 117,542 frames from 11,779 videos of 3531 subjects. The IJB-C verification protocol provides a list of 19,557 genuine comparisons and 15,638,932 impostor comparisons. The result on IJB-C and IJB-B is reported in terms of true accepted rates (TAR) at false accepted rates (FAR) (as defined in \cite{DBLP:conf/cvpr/WhitelamTBMAMKJ17}) equal to 1e-4 to provide a comparable result with the previous works evaluated on these datasets. The achieved verification performances on IJB-B and IJB-C by our MixFaceNet models are reported as part of the 
Table \ref{tab:megaface}. Our MixFaceNet-M and ShuffleMixFaceNet-M models achieved close results to the top-ranked models using significantly fewer FLOPs.

\vspace{-3mm}

\subsection{ Result on MegaFace}
\vspace{-1mm}
The evaluation protocol of MegaFace includes gallery (1m images from Flickr) and probe (FaceScrub and FGNe) sets. In this paper, we use MegaFace \cite{DBLP:conf/cvpr/Kemelmacher-Shlizerman16}  as a gallery set and FaceScrub \cite{DBLP:conf/icip/NgW14} as the probe set to provide a comparable result with the previous works evaluated on this dataset. The MegaFace \cite{DBLP:conf/cvpr/Kemelmacher-Shlizerman16}  contains 1m images of 690K different identities and the FaceScrub contains 100K images of 530 identities \cite{DBLP:conf/icip/NgW14}. The result on MegaFace is reported as identification (Rank-1) and verification (TAR at FAR=1e–6) to be compatible with the previous works evaluated on this dataset \cite{martinez2021benchmarking}. Also, we report the result on the refined version of the MegaFace presented in \cite{deng2019arcface}. The face verification and identification results on the MegaFace and the refined version of MegaFace (noted as MegaFace (R)) are presented in Table \ref{tab:megaface}. For all evaluated models that  have computational complexity between 500 and 1000M FLOPs, our MixFaceNet-M outperformed ProxylessFaceNAS \cite{martinez2021benchmarking}, VarGFaceNet \cite{martinez2021benchmarking}, MobileFaceNetV1 \cite{martinez2021benchmarking}, and ShuffleFaceNet \cite{martinez2021benchmarking,DBLP:conf/iccvw/Martinez-DiazLV19}. And it achieved very close verification and identification results to the top-ranked models- AirFace \cite{DBLP:conf/iccvw/LiWHL19} and MobileFaceNet \cite{martinez2021benchmarking} using less than half the number of FLOPs. 
Also, when the considered computational cost is less than 500M FLOPs, in the third section of Table \ref{tab:megaface}, our ShuffleMixFaceNet-S achieved the highest verification and identification performances.

\vspace{-1mm}
\subsection{Performance vs. Computational Complexity}
\vspace{-1mm}
To present the achieved results in terms of the trade-off between the verification performance and the computation complexity (represented by the number of FLOPs), we plot the number of FLOPs vs. the verification performance of our MixFaceNets and the SOTA solutions.
The plots for the comparisons on the LFW, AgeDB-30, MegaFace, IJB-B, IJB-C and MegaFace(R) benchmarks are presented in Figures \ref{fig:flops} (a), (b), (c) ,(d), (e) and (f). Each of the reported solutions is presented by an indicator on the plot, where an ideal solution will tend to be placed on the top left corner (high performance and low complexity). In most ranges of the number of FLOPs, and on the 6 benchmarks, different versions of our MixFaceNets achieved the highest verification performance. 
Similar conclusions can be made by analyzing the presented values in Table \ref{tab:megaface}.

\vspace{-2mm}
\section{Conclusion}
\label{sec:con}
\vspace{-1mm}

We presented in this paper accurate and extremely efficient face recognition models, MixFaceNets.
We conducted extensive experiments on popular, publicly available, datasets including LFW, AgeDB-30, MegaFace, IJB-B, and IJB-C. The overall evaluation results demonstrate the effectiveness of our proposed MixFaceNets for applications associated with low computational complexity requirements. Our MixFaceNet-S and ShuffleMixFaceNet-S outperformed MobileFaceNets \cite{DBLP:conf/ccbr/ChenLGH18} under the same level of computation complexity ($\leq$500M FLOPs). Also, our MixFaceNet-M is showed to be one of the top-ranked performing models, while using significantly fewer FLOPs than the SOTA models.

\vspace{-1mm}
\paragraph{Acknowledgment:}
This research work has been funded by the German Federal Ministry of Education and Research and the Hessen State Ministry for Higher Education, Research and the Arts within their joint support of the National Research Center for Applied Cybersecurity ATHENE.

\vspace{-2mm}
{\small
\bibliographystyle{ieee}
\bibliography{egbib}
}

\end{document}


\title{MixFaceNets: Extremely Efficient Face Recognition Networks \\ Supplementary Material }
\author{First Author\\
Institution1\\
Institution1 address\\
{\tt\small firstauthor@i1.org}
\and
Second Author\\
Institution2\\
First line of institution2 address\\
{\tt\small secondauthor@i2.org}
}

\maketitle
\thispagestyle{empty}

This supplementary document provides an extended experiments evaluation of our MixFaceNets using different margin-based training losses. 
The investigated loss functions are ArcFace  \cite{deng2019arcface}, CosFace \cite{DBLP:conf/cvpr/WangWZJGZL018} and Combined margin loss \cite{deng2019arcface}.  We use margin value of 0.5 for ArcFace and 0.4 for CosFace. For Combined margin, we set m2 to 0.3 (ArcFace margin) and m3 to 0.2 (CosFace margin). 
We trained all MixFaceNets using the same training settings i.e. number of epochs, model optimizer, learning rate, and batch size as described in the main paper. The description of the datasets evaluation protocol and metric are provided in the main paper.
The following tables are provides:
\begin{itemize}
    \item Table \ref{tab:sup_lfw} shows the achieved accuracies on LFW \cite{LFWTech} and AgeDB-30 \cite{DBLP:conf/cvpr/MoschoglouPSDKZ17}  by MixFaceNets trained with different loss functions. 
    \item Table \ref{tab:sup_ijb} shows the achieved verfication performance on  IJB-B \cite{DBLP:conf/cvpr/WhitelamTBMAMKJ17} and IJB-C \cite{DBLP:conf/icb/MazeADKMO0NACG18} benchmarks.
    \item Table \ref{tab:sup_megaface} presents the achieved verification and identification performances on MegaFace \cite{DBLP:conf/cvpr/Kemelmacher-Shlizerman16} and MegaFace (R) \cite{deng2019arcface}.
\end{itemize}
It can be noticed from the achieved results in Table \ref{tab:sup_ijb}, \ref{tab:sup_ijb}, and \ref{tab:sup_megaface} that the loss functions, we used, achieved very close performances. In general, ArcFace loss achieved slightly higher performance than CosFace and combined margin.

\begin{table*}[]
\centering
\begin{tabular}{|c|c|c|c|c|c|}
\hline
\multirow{2}{*}{Method} &
  \multirow{2}{*}{MFLOPs} &
  \multirow{2}{*}{Params (M)} &
  \multirow{2}{*}{Loss} &
  \multirow{2}{*}{LFW (\%)} &
  \multirow{2}{*}{AgeDb-30 (\%)} \\
 &  &  &             &                &                \\ \hline
\multirow{3}{*}{MixFaceNet-XS} &
  \multirow{3}{*}{161.9} &
  \multirow{3}{*}{1.04} &
  ArcFace &
  \textbf{99.6} &
  \textbf{95.85} \\
 &  &  & CosFace     & 99.48          & 95.32          \\
 &  &  & CombineLoss & 99.533         & 95.47          \\ \hline
\multirow{3}{*}{MixFaceNet-S} &
  \multirow{3}{*}{451.7} &
  \multirow{3}{*}{3.07} &
  ArcFace &
  \textbf{99.65} &
  96.63 \\
 &  &  & CosFace     & \textbf{99.65} & 96.43          \\
 &  &  & CombineLoss & 99.62          & \textbf{96.75} \\ \hline
\multirow{3}{*}{MixFaceNet-M} &
  \multirow{3}{*}{626.1} &
  \multirow{3}{*}{3.95} &
  ArcFace &
  99.68 &
  97.05 \\
 &  &  & CosFace     & 99.68          & \textbf{97.25} \\
 &  &  & CombineLoss & \textbf{99.72} & 96.98          \\ \hline
\end{tabular}%

\caption{MixFaceNets verification accuracies on LFW and AgeDB-30 datasets.}
\label{tab:sup_lfw}
\end{table*}

\begin{table*}[]
\centering
\resizebox{\textwidth}{!}{%
\begin{tabular}{|c|c|c|c|c|c|c|c|}
\hline
\multirow{2}{*}{Method} &
  \multirow{2}{*}{MFLOPs} &
  \multirow{2}{*}{Params (M)} &
  \multirow{2}{*}{Loss} &
  \multicolumn{2}{c|}{MegaFace} &
  \multicolumn{2}{c|}{MegaFace (R)} \\ \cline{5-8} 
 &  &  &             & Rank-1         & TAR at FAR1e–6 & Rank-1 (R)     & TAR at FAR1e–6 (R) \\ \hline
\multirow{3}{*}{MixFaceNet-XS} &
  \multirow{3}{*}{161.9} &
  \multirow{3}{*}{1.04} &
  ArcFace &
  \textbf{74.18} &
  \textbf{89.40} &
  \textbf{89.35} &
  \textbf{91.04} \\
 &  &  & CosFace     & 73.88          & 89.09          & 88.99          & 90.80              \\
 &  &  & CombineLoss & 73.78          & 88.37          & 88.47          & 90.12              \\ \hline
\multirow{3}{*}{MixFaceNet-S} &
  \multirow{3}{*}{451.7} &
  \multirow{3}{*}{3.07} &
  ArcFace &
  76.49 &
  92.23 &
  92.67 &
  93.79 \\
 &  &  & CosFace     & 76.77          & 92.17          & 92.50          & 93.86              \\
 &  &  & CombineLoss & \textbf{77.79} & \textbf{93.67} & \textbf{92.98} & \textbf{94.21}     \\ \hline
\multirow{3}{*}{MixFaceNet-M} &
  \multirow{3}{*}{626.1} &
  \multirow{3}{*}{3.95} &
  ArcFace &
  78.15 &
  \textbf{94.25} &
  \textbf{94.95} &
  \textbf{95.83} \\
 &  &  & CosFace     & \textbf{78.39} & 94.13          & 94.84          & 95.81              \\
 &  &  & CombineLoss & 78.25          & 93.60          & 94.64          & 95.22              \\ \hline
\end{tabular}%
}
\caption{Face identification (Rank-1 \%) and verification (TAR at FAR1e–6) on MegaFace and MegaFace(R).}
\label{tab:sup_megaface}
\end{table*}

\begin{table*}[]
\centering
\begin{tabular}{|c|c|c|c|c|c|}
\hline
\multirow{2}{*}{Method}        & \multirow{2}{*}{MFLOPs} & \multirow{2}{*}{Params (M)} & \multirow{2}{*}{Loss} & \multicolumn{2}{c|}{IJB (TAR at FAR1e–4)} \\ \cline{5-6} 
 &  &  &             & IJB-B          & IJB-C          \\ \hline
\multirow{3}{*}{MixFaceNet-XS} & \multirow{3}{*}{161.9}  & \multirow{3}{*}{1.04}       & ArcFace               & \textbf{88.48}      & \textbf{90.73}      \\
 &  &  & CosFace     & 87.88          & 90.37          \\
 &  &  & CombineLoss & 87.83          & 90.33          \\ \hline
\multirow{3}{*}{MixFaceNet-S}  & \multirow{3}{*}{451.7}  & \multirow{3}{*}{3.07}       & ArcFace               & 90.17               & 92.30               \\
 &  &  & CosFace     & 90.19          & \textbf{92.51} \\
 &  &  & CombineLoss & \textbf{90.24} & 92.19          \\ \hline
\multirow{3}{*}{MixFaceNet-M}  & \multirow{3}{*}{626.1}  & \multirow{3}{*}{3.95}       & ArcFace               & \textbf{91.55}      & 93.42               \\
 &  &  & CosFace     & 91.34          & 93.49          \\
 &  &  & CombineLoss & 91.46          & \textbf{93.60} \\ \hline
\end{tabular}%
\caption{1:1 verification TAR (at FAR=1e-4) on IJB-B and IJB-C.}
\label{tab:sup_ijb}
\end{table*}

\vspace{-1mm}
{\small
\bibliographystyle{ieee}
\bibliography{egbib}
}